\documentclass[12pt,twocolumn]{article}
\usepackage{blindtext}
\usepackage{geometry}[showframe=false,lmargin=1.75in,rmargin=0in,a4paper,width=150mm,top=25mm,bottom=25mm]
\usepackage{graphicx}
\graphicspath{ {./images/} }
\usepackage{amsmath, amsthm, amssymb}
\usepackage{setspace}
\theoremstyle{plain}

\theoremstyle{definition}

\theoremstyle{remark}

\usepackage{amsmath}
\usepackage[utf8]{inputenc}

\author{
  Safa Zaid
  \and
  Aswah Malik
  \and
  Fatima Kisa
  \and
  {National University of Computing and Emerging Sciences (ISB)}
}
\date{}
\title{Jewelry Shop Conversational Chatbot}

\begin{document}
\maketitle
    
\begin{abstract}
Since the advent of chatbots in the commercial sector, they have been widely employed in the customer service department. Typically, these commercial chatbots are retrieval-based, so they are unable to respond to queries absent in the provided dataset. On the contrary, generative chatbots try to create the most appropriate response, but are mostly unable to create a smooth flow in the customer-bot dialog. Since the client has few options left for continuing after receiving a response, the dialog becomes short. Through our work, we try to maximize the intelligence of a simple conversational agent so it can answer unseen queries, and generate follow-up questions or remarks.
 
We have built a chatbot for a jewelry shop that finds the underlying objective of the customer's query by finding similarity of the input to patterns in the corpus. Our system features an audio input interface for clients, so they may speak to it in natural language. After converting the audio to text, we trained the model to extract the intent of the query, to find an appropriate response and to speak to the client in a natural human voice. To gauge the system's performance, we used performance metrics such as Recall, Precision and F1 score.\\\\
\end{abstract}
{\fontfamily{qtm}\selectfont
}

\begin{figure}[ht]
    \includegraphics[width = 230pt]{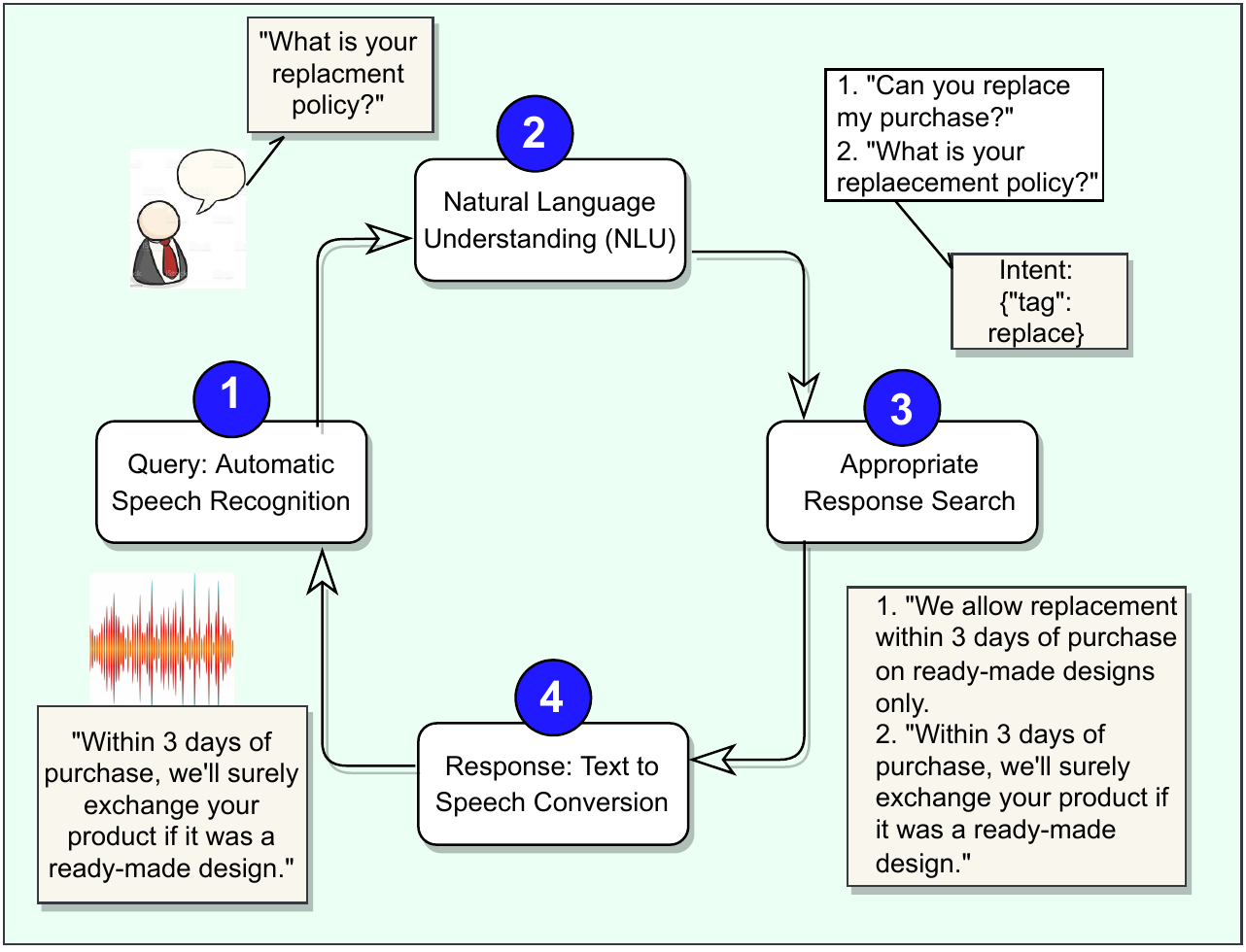}
    \caption{Workflow}
    \label{workflow}
\end{figure}

\section{Problem statement}
Chatbots are increasingly been used in customer service departments since their introduction into the business sector, especially for online businesses. Unfortunately, basic task-oriented chatbots are unable to address client enquiries that are not present in their Frequently Asked Questions (FAQ) dataset. We will employ state-of-the-art Natural Language Processing (NLP) technology to construct a conversational chatbot that can speak more robustly with its clients. It will be capable of responding to a wider range of queries and presenting the customer with occasional prompts. This will make the customer-chatbot conversation more natural and will promote brand loyalty.

\section{Introduction}
\begin{enumerate}
    \item {\em Problem Details}
    Jewelry shops are rarely open 24/7, thus their selling time is constrained. For an online jewelry shop, though its website is accessible at any time  of the day from across the globe, it is not feasible for the shop to satisfy individual queries of each of its clients. The employment of a chatbot on the jewelry shop's website will allow customers across the globe to enquire about the shop at any time of the day. This better customer service will help retain clients and increase sales. Most chatbots that we interact on websites can answer only a given set of queries since they are rule-based chatbots. This means that if a query does not exactly match a previously saved pattern in the model's corpus, the bot would be unable to respond to it. Using NLP and Machine Learning (ML) models, we developed a conversational chatbot which not only resolves customer issues but also generates follow-up questions and remarks, making the conversation more human-like for the customer. \cite{14}
    
    The image below shows how a Rule-based and AI-based chatbot are different.

\begin{figure}[ht]
    \includegraphics[width = 250pt]{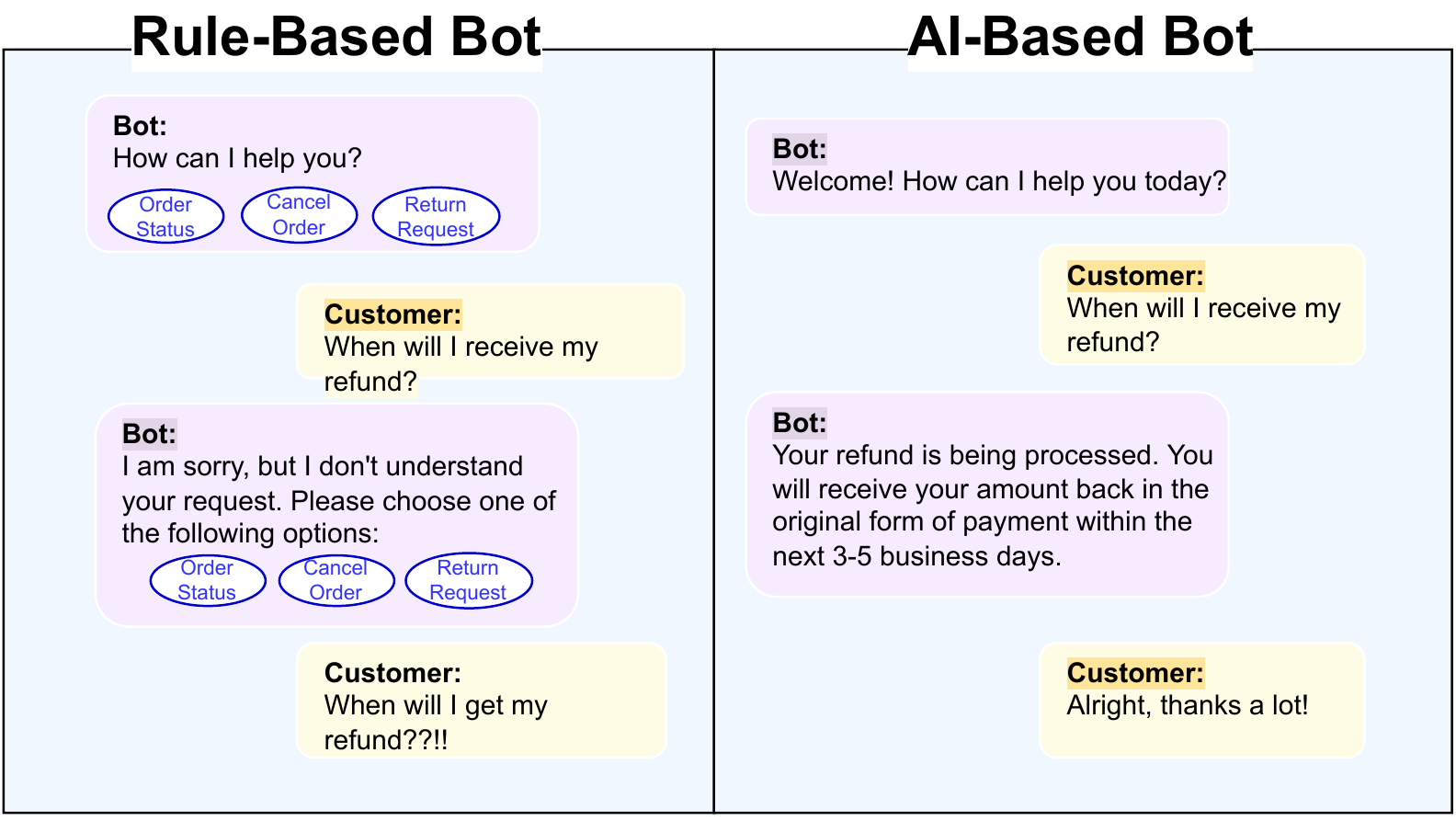}
    \caption{Rule-based vs. AI-based Chatbots}
    \label{rule_ai_based}
\end{figure}

    \item {\em Motivation}
    Buying and selling items or services through text-messaging applications is part of conversational commerce. Companies are putting a lot of money into digitizing customer service via social media and company websites. They hope to create a more personalized sales experience and stay competitive in the market by these means. A competent sales chatbot will not only respond to consumer questions, but will also provide product recommendations based on the user's apparent preferences, imitating the work of a salesman in a physical store. Conversational AI, which includes speech recognition, sentiment and semantic analysis, and context-based response creation, is found to assist in creating a personalized customer experience. 52 percent of organizations indicated they increased their usage of automation and conversational interfaces following COVID-19, and 86 percent of respondents said AI has become "mainstream technology" in their company \cite{6}.

    \item {\em Background}
    
     A dialog system or conversational agent communicates with users in natural language, that is text and/or speech. They can be divided into two classes:
     
    \begin{enumerate}
    \item {\em Task-oriented dialog agents}: These perform the basic purpose of following the given directions or answering questions on corporate websites. 
    \item {\em Chatbots}: These are designed for extended, unstructured and sometimes even multi-contextual conversations. They can be used both for entertainment and for making the interactions of task-oriented agents more natural.
    \end{enumerate}
    
     There are two basic types of chatbots- Rule-based and Artificially Intelligent(AI) or Corpus-based chatbots. The first actual chatbot was rule-based . Rule-based models are simpler to implement but have limited capabilities. They answer queries by pattern matching and thus, can often produce faulty or no solutions when the user query does not match with any recognized pattern. Contrarily, AI models are primarily based on machine learning algorithms which use existing corpora of human conversations to train them. Unlike Rule-based models, AI-based models can understand the user intent and context, and over time, use negative feedback on their mistakes to improve performance.
     
     Within AI-based chatbots, there are two further sub-types, namely Information-Retrieval(IR) chatbots and Generative chatbots. Information Retrieval models are trained on a textual dataset, primarily designed to retrieve the information based on user input. The knowledge base for this type of model is usually formed using a database of query-answer pairs. When the person queries the chatbot, the model finds similarities in the query and the chat index.
     
     Generative Models generate entirely new sentences based on the user queries. However, they need to be trained on a large dataset of phrases and real conversations. The model learns sentence structure, syntax, and vocabulary with the aim of generating linguistically correct and contextually appropriate answers.
     
     Neural Networks (NN), first introduced in the late 1980’s are large computational networks that are trained on large datasets in order to approximate some complex target function. They are computational systems that try to solve problems like a human brain, and hence can be used to solve problems like natural language understanding, intent classification and question answering.

\end{enumerate}

\section{Related work}
    Digital commerce has resulted in customers demanding round-the-clock customer service by businesses. Due to this, chatbots are increasing in popularity among businesses and consumers alike. More and more companies are ready to pay high amounts of money for the development of these chatbots. As chatbots raise customer engagement via messaging, text, or speech, they are deployed on social and work platforms such as Facebook Messenger, WhatsApp, WeChat, and Slack.
    
    Our chatbot is inspired by many chatbots that we have around ourselves. Early conversational systems like ELIZA \cite{7} (in 1966) and ALICE \cite{8} (in 1995), which were rule-based and had a constrained scope, held the purpose to mimic human-human text-based conversation. However, the rules were hand-written and responses were generated by keyword pattern matching \cite{9}.
    
    In 2000, another major dialog system was introduced, called the DARPA communicator program \cite{10}. Its key features were goal-oriented natural language understanding of requests, conducting dialog and performing tasks. Further, this chatbot had a Learning-based model that used statistical models for understanding spoken inputs in addition to textual inputs. However, its biggest technical limitation was that its performance was poor outside of its well-defined domains. 
    
    In 2011, Siri \cite{11}, the first widely deployed learning-based Intelligent Personal Assistant (IPA), was developed with an open domain using a Deep Neural Network to convert acoustic patterns in the input voice to form a probability distribution over speech sounds. Like other IPAs, Siri provides both reactive assistance -like generating weather reports- and proactive assistance -like reminding of a friend’s birthday- to users so that they could accomplish a variety of tasks. However, it lacks emotional engagement with its users.
    
    The first widely deployed social chatbot, XiaoIce \cite{12} was designed in 2014 and is used to date. In addition to assisting users in various tasks, it has its own personality and has the ability to create emotional attachment with 3 its users using Emotional Intelligence learning based models in an open-domain using text, speech and images. However, it often shows inconsistent responses and personality traits in long conversations. 
    
    In previous years, Sequence-to-Sequence models\cite{13}, a special class of RNNS, were used for obtaining valuable results after training on open-domain knowledge. They can also be integrated with other algorithms for domain-specific analyses. Nonetheless, the major drawback of these models is that the entire information (including the past context) of the input sentence into fixed length context vector. Thus, as the sentence or context gets longer, more information is lost and the model responds with decreasing coherence.


\section{Methodology}

     Our chatbot has an audio input interface for the customers, meaning the customers can speak to the chatbot in natural language. This audio is converted to text by Python's Speech Recognition library, SpeechRecognizer \cite{1}. This text is then associated to certain fixed intent in the corpus. Against each intent, we have multiple equivalent responses. Thus, after the customer pattern has been classified as belonging to a particular intent, a seemingly random response is generated. Thereafter, even if the customer asks the same question repeatedly, the response generated is very likely to vary, as well. Furthermore, the chatbot occasionally presents the customer with a followup remark or question to imitate the human conversation. To give our human-chatbot conversation a more natural touch, the chatbot also speaks to the customer in the voice of a man. For this feature, we used the Python library, pyttsx3\cite{2}. The chatbot will continue the conversation with the customer until it classifies an input pattern as a “goodbye” intent. In the case that a “goodbye” pattern is found, the chatbot greets its client appropriately and ends the conversation.
     
     We developed the chatbot in three different ways:
    
    \begin{enumerate}
    \item
            For our first method, we built the chatbot based on TensorFlow's Keras Sequential model -a feed forward multi-layer neural network. The customer’s input query is pre-processed and compared with the template “patterns” or “queries” in our self-generated customer service dataset. The pre-processing steps include tokenizing, stemming, lemmatization and removing punctuation from our dataset. The input and output layers of the Neural Network consist of One-Hot-Encoded (OHE) embeddings to describe patterns and predicted intents respectively. During the model's feed forward pass, it optimizes the layer weights using Stochastic Gradient Descent (SGD) and has a standard learning rate of 0.01. The model uses the Rectified Linear Unit (ReLu) as the activation function between outputs and inputs of adjacent hidden layers. At the last layer, Softmax is applied to our multinomial linear regression model to normalize the output layer results.\\
    \item
            In the second method, the embeddings from One-Hot-Encoding were replaced by embeddings generated by SentenceTransformer model. This was done to observe how naive One-Hot-Encoded embeddings and the more meaningful SentenceTransformer embeddings of size 384x1 would affect the classifier model's predictions.\\
    \item
            In the last method, the SentenceTransformer model from the previous variation was used, but the Intent Classification Model was replaced with a Cosine Similarity Function. This function determined the pattern from the corpus to which the input customer query is most similar. The intent of the matched pattern is extended to the input and the query is assigned its tag. Finally, a response and optional followup is generated as mentioned above. Following is an example of the final method's working:\\\\

        {\fontfamily{qtm}\selectfont
            {\small Input Query: What time can I visit your shop?\\
                Matched Pattern: What are your shop timings?\\
                Predicted Intent: Timing\\
                Response: Our shop opens at 8 am and closes at 11 pm.\\
                Follow up: We are open for the longest hours in the market!\\
                }
            }
    \end{enumerate}

\section{Evaluation and Experiments}

Our chatbot was built with features many chatbots do not contain. For example, most bots take input and produce output as text, which is not how humans naturally communicate. To avoid a tedious conversation, the chatbot is enabled with the feature of Speech Recognition. However, a clear voice and quite environment is required for ensuring an appropriate output.

We applied stemming on the user input to easily identify different forms of a word having a similar effect on intent classification. We tested this feature by saying different sentences with different sentence structures but same vocabulary to check if the bot intelligently finds the stem word and responds correctly. For example:\\\\

    {\fontfamily{qtm}\selectfont
        {\small
        Input Query 1: What time can I visit your shop?\\
        Stemmed Query 1: What time can I visit your shop?\\
        Input Query 2: When is your shop open for visiting?\\
        Stemmed Query 2: When is your shop open for visit ?\\
        Pattern for Queries 1 \& 2: What are your shop timings?\\
        Intent for Queries 1 \& 2: Timing\\
        }
    }

An interesting feature of our chatbot is that it does not produce the same response on a repeated query. For this, we have used the random.shuffle() utility from Python's random library on the list of responses in our corpus. In addition, it sometimes asks the customer follow up questions for their better understanding unlike other chatbots, which and never initiate the conversation themselves.

First of all, the experiment we conducted was to give same input again and again to confirm that our chatbot always gives a different answer to same input considering that the customer didn't understand its previous response as demonstrated in example below.

{\fontfamily{qtm}\selectfont
    {\small
    Query 1: What time can I visit your shop?\\
    Response 1: Our shop opens at 8 am and closes at 11 pm.\\
    
    Query 2: What time can I visit your shop?\\
    Response 2: You can come anytime between 8 am and 11 pm!\\
    }
}

We evaluated our chatbot using inputs from different intents to calculate its F1 score using a Confusion Matrix for each of the three implementations.


\begin{table}[ht]
\centering
 \begin{tabular}{p{0.25\textwidth}p{0.20\textwidth}} 
 \hline
 Implementation & F1 Score \\ [0.5ex] 
 \hline\hline
 OHE with NN & 0.592 \\
 \hline
 Sentence Embedding with NN  & 0.649 \\
 \hline
 Sentence Embedding with Cosine Similarity & 0.852 \\ [1ex] 
 \hline
 \end{tabular}
\end{table}

As can be seen from the table above, the One Hot Encoding generated naive and somewhat meaningless embeddings for the Neural Network classifier. Further, since the dataset on which the Classification Model was trained, was built by only three people, its limited size adversely affected the training of the NN. In comparison, the sentence transformer embeddings made the input to the classifier clearer as the embedding was more meaningful and its vector was larger. As for the implementation with the sentence embedding paired with the Cosine Similarity function, the results were the best, as this function was not affected by the corpus's size like the NN.

\section{Future Work}
Although our system works well for most customer queries, the knowledge domain of the chatbot is limited due to small dataset size. Its size can be expanded by adding more intents, patterns and responses. In addition, run-time calculations for price of a set could be an added feature to our bot. Further, run-time scraping could be enabled to answer queries not present in the dataset. Lastly, these unknown intents could be dynamically inserted into the corpus to reduce the number of scrapings required in an unseen scenario.

\bibliographystyle{plain}


\end{document}